%% file: main.tex
\documentclass[sigconf]{acmart}
\usepackage{pgfplots}
\usepackage{multirow}
\usepackage[algo2e,ruled,vlined, noend, linesnumbered]{algorithm2e} 
\usepackage{mathtools, amsfonts}
\usepackage{caption}
\usepackage{xcolor}
\usepackage{tikz}
\usepackage{tikz}
\usetikzlibrary{shadows}
\usetikzlibrary{arrows}
\usepackage{graphicx,subcaption}
\usepackage{tcolorbox}

\setcopyright{rightsretained}
\begin{document}
\copyrightyear{2021}
\acmYear{2021} 
\acmConference[KDD '21]{Proceedings of the 27th ACM SIGKDD Conference on Knowledge Discovery and Data Mining}{August 14--18,2021}{Virtual Event, Singapore}
\acmBooktitle{Proceedings of the 27th ACM SIGKDD Conference on Knowledge Discovery and Data Mining (KDD '21), August 14--18, 2021, Virtual Event, Singapore}
\acmDOI{10.1145/3447548.3467381}
\acmISBN{978-1-4503-8332-5/21/08}
\fancyhead{}

\title{Spectral Clustering of Attributed Multi-relational Graphs}


\author{Ylli Sadikaj}
\affiliation{%
\department{Faculty of Computer Science}
  \institution{University of Vienna}
  \streetaddress{Waahringer Straße 29}
  \city{Vienna}
  \country{Austria}
}
\email{ylli.sadikaj@univie.ac.at}

\author{Yllka Velaj}
\affiliation{%
  \department{Faculty of Computer Science}
  \institution{University of Vienna}
  \streetaddress{Währinger Straße 29}
  \city{Vienna}
  \country{Austria}
  }
\email{yllka.velaj@univie.ac.at}

\author{Sahar Behzadi}
\affiliation{%
\department{Faculty of Computer Science}
  \institution{University of Vienna}
  \streetaddress{Währinger Straße 29}
  \city{Vienna}
  \country{Austria}
  }
\email{sahar.behzadi@univie.ac.at}

\author{Claudia Plant}
\affiliation{%
\department{Faculty of Computer Science, ds:UniVie}
  \institution{University of Vienna}
  \streetaddress{Währinger Straße 29}
  \city{Vienna}
  \country{Austria}
  }
\email{claudia.plant@univie.ac.at}

\renewcommand{\shortauthors}{Y. Sadikaj et al.}

\begin{abstract}
Graph clustering aims at discovering a natural grouping of the nodes such that similar nodes are assigned to a common cluster. Many different algorithms have been proposed in the literature: for simple graphs, for graphs with attributes associated to nodes, and for graphs where edges represent different types of relations among nodes. 
However, complex data in many domains can be represented as both attributed and multi-relational networks.

In this paper, we propose SpectralMix, a joint dimensionality reduction technique for multi-relational graphs with categorical node attributes. SpectralMix integrates all information available from the attributes, the different types of relations, and the graph structure to enable a sound interpretation of the clustering results.
Moreover, it generalizes existing techniques: it reduces to spectral embedding and clustering when only applied to a single graph and to homogeneity analysis when applied to categorical data. 

Experiments conducted on several real-world datasets enable us to detect dependencies between graph structure and categorical attributes, moreover, they exhibit the superiority of SpectralMix over existing methods.
\end{abstract}



\maketitle

\section{Introduction}
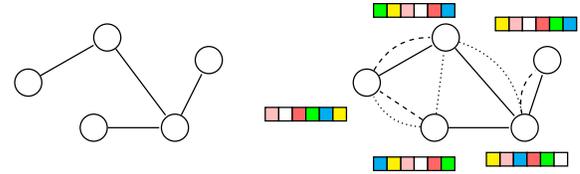
\begin{figure}[t]
	\centering
	\scalebox{0.6}{\input{img/fig1.tex}}
	\caption{\footnotesize{Example of a simple graph (left) and an attributed multi-relational graph (right). Different types of lines between nodes represent different relation types.}}\label{fig:ex}
\end{figure}

Complex data in many domains including biology, neuroscience, social and collaboration networks etc.  can be represented as  attributed multi-relational graphs. The nodes are often characterized by many attributes and the links can be of different types, as depicted in Figure~\ref{fig:ex}. In social networks, for instance, nodes represent users who are characterized by attributes like gender, age, hometown, hobbies etc., and edges describe the different types of links established by interaction on different platforms like Facebook and Twitter. In the field of neuroscience, the data obtained by neuroimaging techniques can be modelled using nodes for the anatomical brain regions, attributes for the region size and type of tissue, and edges for the structural and functional connectivity links, as measured by structural and functional Magnetic Resonance Imaging (MRI and fMRI)~\cite{Meier2016AMB}. 

Despite the importance of attributed multi-relational graphs in modeling many real-world domains, they have received little attention, in particular in the task of clustering.

Clustering is one of the main research directions in data mining tasks. It aims at discovering a natural grouping of the data such that similar data objects are assigned to a common cluster. Clustering is very helpful in order to obtain an understanding of the major patterns in large datasets like attributed multi-relational networks.
Existing approaches are mainly designed for simple graph clustering~\cite{PICS,Inc_Cluster, SA_Cluster, GBAGC, CESNA}, for fuzzy clustering~\cite{CLAMP} that identifies overlapping clusters in which an object belongs with a membership probability, for clustering of networks with numerical attributed nodes~\cite{HASCOP}.

Attributes and multiple types of relationships between nodes of a graph can yield complementary information. This information is valuable for detecting clusters of nodes with similar characteristics and can be used in different applications, e.g., targeting ads. 
Thus, we develop SpectralMix, a novel algorithm that provides a comprehensive approach to joint dimensionality reduction and spectral clustering of attributed multi-relational graphs  with  categorical  node  attributes. Given a multi-relational attributed graph $G$ and the desired dimensionality $d$, SpectralMix defines a mapping $\Phi: G \rightarrow \mathbb{R}^d $ that minimizes the distance between nodes with similar attributes. The joint low-dimensional vector space generated by SpectralMix reduces noise, emphasizes the major patterns in the data, and enables subsequent data mining. 
By application of the $K$-means algorithm we obtain a clustering inspired by spectral clustering methods. But we could also perform outlier detection or visualize the data when we select a 2D or 3D vector space. 
Moreover, other very important aspects of SpectralMix are that: there is only a single dimensionality parameter $d$ that needs tuning, and SpectralMix generalises other existing techniques. When considering the setting of a single input graph without any node attributes as input data, our mapping is the same as Laplacian Eigenmaps~\cite{DBLP:conf/nips/BelkinN01}. When considering only categorical data without relational information, our mapping coincides with Homogeneity Analysis~\cite{JSSv031i04}, a technique from statistics for categorical PCA~\cite{Jolliffe1986}. 

To assess its performance, we evaluate SpectralMix based on   clustering and data visualization tasks. The experiments conducted on several real-world networks 
enable us to detect dependencies between graph structure and the categorical attributes,  moreover,  they exhibit the superiority of SpectralMix over alternative baselines.

\section{Related works}\label{sec:rel_works}
Recently, the problem of clustering graphs with node attributes has received a lot of attention. 
The main goal in this field has been to improve the quality of the clustering by combining both structural and attribute properties. 

While it may seem a natural extension, many methods~\cite{ GBAGC, SSCG, BAGC} cannot be directly applied to attributed multi-relational graphs. They propose algorithms for detecting densely connected components~\cite{Inc_Cluster, SA_Cluster, GBAGC, CESNA}, or for identifying clusters of objects having similar connectivity~\cite{PICS, HASCOP} but they either ignore the multiple types of edges~\cite{UNCut, NNM, DBLP:journals/isci/0007YLDPB21} and/or deal with only one attribute type~\cite{CAMIR}. 
A technique, that has been used frequently to overcome these problems, is to project the attributed multi-relational graph to a single weighted graph where weights represent a combination of attribute and structural similarities, and then apply a clustering algorithm for weighted graphs with the aim of performing clustering and graph outlier detection~\cite{FocusCO} or identify strong community structures~\cite{Steinhaeuser98}. This projection, however, causes information loss and, as a consequence, it limits the clustering accuracy. Similarly, graph embeddings techniques have been designed mainly for single graphs such as FATNet~\cite{DBLP:journals/isci/0007YLDPB21} which integrates the global relations of the topology and attributes into robust representations.

Different works have presented results on spectral clustering on  multi-layer networks~\cite{Spectral1, Spectral2, Spectral5}, while very few, to the best of our knowledge, investigate spectral clustering methods on attributed graphs~\cite{Spectral4}. 
Whereas, there are methods not designed for graphs, which exploit the fusion of the cluster-separation information from all eigenvectors to achieve a better clustering~\cite{DBLP:conf/kdd/YeGPB16} or perform clustering of numerical and categorical data based on homogeneity analysis~\cite{DBLP:conf/kdd/Plant12}.

Recently, many works in the area of graph embedding and clustering have been published and they can be categorized in the three groups: a) methods for single graphs with attributes, ANRL~\cite{ijcai2018-438} and DGI~\cite{velickovic2018deep}; b) methods for multi-relational graphs without attributes, DMGC~\cite{luo2020deep} and CrossMNA~\cite{10.1145/3308558.3313499}; and c) methods for attributed multi-relational graphs, HAN~\cite{han2019}, MARINE~\cite{10.1145/3308558.3313715}, DMGI~\cite{park2019unsupervised}.

ANRL~\cite{ijcai2018-438} is a deep neural network model designed for attributed networks representation learning which integrates network structural proximity and node attributes affinity into low-dimensional representation spaces. 
DGI~\cite{velickovic2018deep} is an approach for learning unsupervised representations on graph-structured data and it relies on maximizing mutual information between patch representations and corresponding high-level summaries of graphs derived using established graph convolutional network architectures. 
DMGC~\cite{luo2020deep} is a deep learning based multi-relational graph clustering method which aims at performing two procedures: graph clustering and cross-graph cluster association.  
CrossMNA~\cite{10.1145/3308558.3313499} is an embedding method that integrates cross-network information to refine two embedding vectors for alignment tasks: inter-vector and intra-vector.
MARINE~\cite{10.1145/3308558.3313715} is an unsupervised embedding method for both simple and multi-relational networks that preserves the proximity and attributes information in addition to various types of relations. 
Furthermore, HAN~\cite{han2019} is a semi-supervised graph neural network framework that can be applied to graphs with various types of nodes and edges. It employs node-level attention and semantic-level attention simultaneously. DMGI~\cite{park2019unsupervised} is an unsupervised network embedding method for attributed multi-relational graphs. It is a modified approach of DGI, and it jointly integrates the embeddings from multiple types of relations between nodes through a consensus regularization framework, and a universal discriminator. 
 
To conclude, the previous methods differ from ours in the following ways: (i) they do not consider the existence of multiple edge types, (ii) they deal with only one type of attributes, (iii) their final node embedding is represented in higher dimensions, 100 or 200 dimensions, (iv) they can perform well only on one graph category: multigraphs with attributes or multigraphs without attributes.

SpectralMix outperforms the above methods because it exhibits good results when applied to graphs with one or many different edge types, to graphs with or without node attributes, and a combination of those. Moreover, the output from our algorithm can be represented in 2D or 3D space and has more meaningful cluster structure.

\section{Notation and Problem Definition}

We consider an undirected node-attributed multi-relational network, (also referred to as multigraph), $G = (V, E, \mathcal{R})$ where  $V = \{v_1, \dots, v_n\}$
represents the set of nodes. Each node $v_i$ is characterized by a set $\mathcal{A}=\{A_1, \dots, A_c\}$ of $c$ categorical attributes, i.e., $|\mathcal{A}|=c$, and $c$ is the dimensionality of the attribute space. 
We denote categorical attributes by upper case letters. Let $A_j$ for $j=1,\dots,c$ be a categorical attribute having $k_j$ distinct values, with $k_j = 2$ for binary attributes. For each categorical attribute, we allow $k_j$ for $j=1,\dots,c$ associated to it, to have a different value. Moreover,  $v_i.A_j$ denotes the value or category of vertex $v_i$ in attribute $A_j$. Thus, the total number of categories, denoted by $C$, of all $c$ categorical attributes is $C=\sum_{j=1}^{c}k_j$. 

We denote by $\mathcal{R}$ the set of different relation types and,  therefore, $|\mathcal{R}|$ denotes the dimensionality of the network.
The set of edges $E$ consists of the tuples $e = (v_i, v_j, w_{ij}, r)$ where $w_{ij}$ represents the weight of the edge of type $r$, and $r\in \mathcal{R}$ is the type of relation.  We denote by $E_r$ set of edges of type $r$, for $r=1, \dots, |\mathcal{R}|$.

Our goal is to define a mapping $\Phi: G \rightarrow \mathbb{R}^d $, where $d$ is the desired dimentionality, such that the distance between connected nodes with similar attributes is minimal. We define by $\mathcal{O}$ and $\mathcal{M}$  the matrices of size $n\times d$ and $C \times d$, respectively, where each entry  $o_{il}$ and $m_{jl}$ for $i=1,\dots, n$, $j=1,\dots, C$ and $l=1,\dots, d$, is a real value denoting the coordinate of the node or the categorical attribute according to the embedding.

Our algorithm, sketched in Algorithm~\ref{fig:algo}, takes as input a multi-relational graph $G$, a categorical attribute matrix of size $n\times c$, and the desired dimensionality $d$ of the embedding. It returns as output a $d$-dimensional feature space representation of data objects $\mathcal{O}$ and categories $\mathcal{M}$, and, if desired, also a clustering of the data objects. Hence, $\mathcal{O}$ and $\mathcal{M}$  are two matrices of size $n\times d$ and $C \times d$, respectively.

Our framework can also be applied to cluster multi-layer graphs where there are inter-layer edges  between any pair of nodes. Given a multi-layer graph, we build a multi-relational graph with the same set of nodes and the same attributes associated to them. For the set of edges, we consider as many relations as there are layers, we also add relations representing the edges from nodes in one layer to the other layers, if those edges exist.

\section{Embedding Multidimensional Attributed Graphs}\label{sec:emb}
Our problem setting is characterized by mixed-type data of different modalities. In particular, we have a set of nodes which are characterized by multiple categorical attributes and linked by different relations. We integrate all the different modalities by a joint vector space representation.
Our objective function for this embedding combines the ideas of spectral embedding of graphs with homogeneity analysis of categorical data. For $n$ nodes we derive low-dimensional coordinates minimizing the following objective function:

\vspace*{-5mm}
\begin{multline}
 \min_{\mathcal{O},\mathcal{M}} \sum_{r = 1}^{|\mathcal{R}|} \alpha_r \cdot \left(\sum_{e=(v_i, v_j, w_{ij}, r) \in E_r}w_{ij} \cdot \sum_{l=1}^{d} ||o_{il} - o_{jl}||^2\right)
 \\+ \sum_{i=1}^{n} \sum_{j = 1}^{|\mathcal{A}|} \alpha_j \cdot \sum_{l=1}^{d}||o_{il} - m_{il}||^2 \\ \mbox{subject to  } \mathcal{O}^T \mathcal{O} = I_n.\label{eq:obj}
\end{multline}

As mentioned in the previous section, $ \mathcal{O}$ is a $n \times d$ matrix of low-dimensional coordinates. Every row represents the coordinate vector for one data object (node). The first summand in Equation~\ref{eq:obj} represents the contribution of the relational part of the data, i.e., of the $|\mathcal{R}|$-dimensional graph on the position of a data object. For every graph we consider all edges between $v_i$ and vertices $v_j$ and determine the coordinate matrix $\mathcal{O}$ such that the squared Euclidean distances are minimized.  We denote by $o_{il}$ the coordinate of the node $v_i$ in the dimension $l$. As in spectral embedding techniques we place the vertices as close as possible to their neighbors, but now not considering a single graph but multiple different relations. 

\begin{figure}[t]
	\centering
	\scalebox{0.55}{\input{img/fig2.tex}}
	\caption{\footnotesize{Example of an attributed multi-relational graph (left) and a bipartite graph for its categories for attribute `Hometown'.}}\label{fig:bip_graph}
\end{figure}
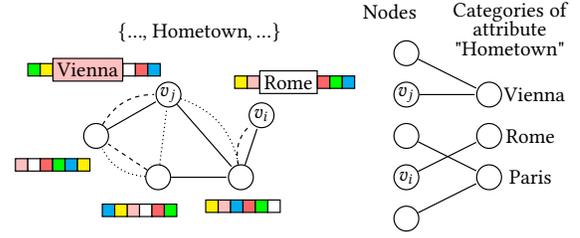 

The second summand of the equation represents the categorical information. Categorical data establishes a bipartite graph between data objects and their categories. Every categorical variable establishes its own bipartite graph, where object $v_i$ is connected to the category or value of its attribute $v_i.A_j$. 
An example of a bipartite graph is shown in Figure~\ref{fig:bip_graph}. 

We now also consider the $c$ categorical graphs in determining the joint vector space embedding. Inspired by Homogeneity Analysis, our algorithm SpectralMix will discover for every category $k_j$ of every categorical attribute $A_j$ a position in $\mathbb{R}^d$. In Equation~\ref{eq:obj} we denote the position of the category of object $v_i$ in dimension $l$ by $m_{il}$. The matrix $\mathcal{M} \in C \times d$ contains these category coordinates as row vectors. 

The coefficients $\alpha_r, \alpha_j$ represent weighting factors for the relations types and the categorical attributes in the input data. It is possible to consider the unweighted objective function setting all to $1$. However, the total number of edges in the multi-relational graph and the edges in the graphs generated by the attributes tend to be very different as the categorical data establishes bipartite graphs where every node is connected to only and exactly one category. As default weighting scheme we, therefore, suggest to set the weight factors such that every modality has the same weight in the low-dimensional representation.

In order to avoid trivial solutions, we require the matrix $\mathcal{O}$ to be column-orthonormal. Without this constraint, a minimum of the objective function would be achieved by mapping all vertices to one common location.

\begin{algorithm2e}[t]
\footnotesize{
	\SetAlgoLined
	\KwIn{ Graph $G$, attribute matrix, dimensionality $d$}
	\KwOut{$d$-dimensional feature space representation of data objects $\mathcal{O}$ and categories $\mathcal{M}$, clusters $clu_1,\dots, clu_k$ }

Initialize matrices $\mathcal{O}$ and $\mathcal{M}$ randomly\;
\Repeat{convergence}{\tcc{update object coordinates}
\For{$i=1, \dots, n$}{
\For{$r=1, \dots, |\mathcal{R}|$}{
\For{$p=1, \dots |N_v|\text{ with } N_v=\{v_p| (v_i, v_p)\in E_r \}$ }{
\For{$l=1, \dots, d$}{$o_{il}\leftarrow o_{il} + \alpha_r\cdot w_{ip}\cdot o_{pl}/sum_i(G)$\;}
}}
\For{$j=1, \dots, C$}{
	\For{$l=1, \dots, d$}{$o_{il}\leftarrow o_{il}+ \alpha_i\cdot m_{jl}/sum_i(A)$\;}
}
}
Apply orthonormalization to ensure $\mathcal{O}^T \mathcal{O}= I_n$\;
\tcc{update category coordinates}
\For{$i=1, \dots, n$}{
	\For{$l=1, \dots, d$}{
		\For{$j=1, \dots, C$}{$m_{jl}\leftarrow m_{jl} + o_{il}/tot_j$\;}
}}
}

\eIf{clustering is desired}{
perform k-Means on the rows of $\mathcal{O}$ with $k = d-1$\;
\textbf{return }clusters $clu_1,\dots, clu_k$\;}
{
\textbf{return }$\mathcal{O}, \mathcal{M}$\;
}
\caption{SpectralMix}\label{fig:algo}}
\end{algorithm2e}

Our algorithm, reported in Algorithm~\ref{fig:algo}, starts by initializing every element $o_{il}$ for $i=1,\dots, n$ and $l=1,\dots, d$, in  $\mathcal{O}$ with a real number selected at random. In this matrix, every row represents the data object associated to a vertex of the graph $G$, and every column is one of the coordinates in the embedding space.

Then, we compute the value of our objective function as described in Equation~\ref{eq:obj}, and at each iteration we update the coordinates in $\mathcal{O}$ and $\mathcal{M}$ in order to obtain a smaller value for our objective. If this value does not decrease after the updates, then the algorithm converges and returns the final node embeddings for the data objects (nodes) and the attribute categories.

In particular, when we update the object coordinates we take into account the type of relations in the graph (lines $4-7$) and the attributes (lines $8-10$). For the graph contribution, we consider, for all the different relation types, the neighbors $v_p$ of node $v_i$ and the weight of these edges. We update each coordinate adding to it the product of the weighting factor $\alpha_r$, the weight $w_{ip}$ of the edge $(v_i, v_p, w_{ip}, r)$, the coordinate of the data object for node $v_p$, and divide by $sum_i(G)$ computed as the sum of the size for the neighborhood of node $v_i$ in relation $r$.
For the attribute contribution, we update each coordinate adding to it the product of the weighting factor $\alpha_i$ for the bipartite graph generated by the category, the coordinate of the category for attribute $A_j$, and divide by $sum_i(A)$ computed as the sum of the weighting factor for category $i$. Then we apply the orthogonalization algorithm.

When we update the coordinates of the categories, we add to each of them the coordinate $o_{il}$ of node $v_i$ and dived by the total number of elements that belong to that category. 
We highlight that our algorithm has only one parameter $d$, desired dimensionality, to be tuned by human, while all the other parameters (weights) can be learned during optimization.

We provide now a proof of convergence of our algorithm.
	\begin{theorem}
		The SpectralMix algorithm converges in a finite number of iterations.
	\end{theorem}
\vspace*{-5mm}
\begin{proof}

		We say that our algorithm converges if the difference in the value of the objective function $\mathcal{F}$ between two consecutive iterations, $t$ and $t+1$, of our algorithm is bounded by a small positive constant $\epsilon$: $\mathcal{F}_{t+1}-\mathcal{F}_t<\epsilon$.
		The idea is to show that the two steps of the algorithm, i.e., the update of the object coordinates and the update of the attributes coordinates, decrease the objective function and that the difference  $\mathcal{F}_{t+1}-\mathcal{F}_t$ tends to zero. 
		Suppose that the algorithm proceeds from iteration $t$ to iteration $t+1$, it suffices to show that \textcolor{red}{:} $$||o_{il} - o_{jl}||^2_t\geq ||o_{il} - o_{jl}||^2_{t+1}\text{ and  }||o_{il} - m_{il}||^2_t\geq ||o_{il} - m_{il}||^2_{t+1}.$$
		In lines $4-7$ of Algorithm~\ref{fig:algo}, the coordinates of a node are updated by taking into account the position of all its neighbors in the graph. Therefore, every node moves towards nodes that are closer in the space, reducing the distance $||o_{il} - o_{jl}||^2$. 
		This observation also holds for the attributes as they are represented as nodes in bipartite graphs, in lines $12-15$ the coordinates are updated so that these attributes moves towards the nodes to which they are associated with, therefore the distance $||o_{il} - m_{il}||^2$ decreases.
		This information is also considered while updating the coordinates of the objects in lines $8-10$.
		As the algorithm updates the coordinates of the nodes, the distance between two nodes $v_i$ and $v_j$ represented by the coordinates $o_{il}$ and $o_{jl}$, decreases but their distance to other neighbors might increase. This does not affect the trend of the objective function. We can see that by distinguishing two cases:
		(a) Edge $e=(v_i, v_j, w_{ij}, r)$, for a given relation $r$, is the edge with highest weight $w_{ij}$ for both $v_i$ and $v_j$. In this case, when the distance between $v_i$ and $v_j$ decreases, due to the update in line $7$, it produces a decrease in the value of $\mathcal{F}$. This difference compensates the distances that may increase from $v_i$ and $v_j$ to any of their neighbors $v_k$ such that $w_{ij}>w_{ik}$ and $w_{ij}>w_{jk}$.  
		(b) Edge $e=(v_i, v_j, w_{ij}, r)$, for a given relation $r$, is the edge with highest weight $w_{ij}$ for $v_i$ but not for $v_j$ that is connected to $v_k$ by an edge of weight $w_{jk}>w_{ij}$. In this case, the distance between $v_i$ and $v_j$ might not decrease, because as $v_i$ moves towards $v_j$, the latter node moves towards $v_k$. Even in this case, the value of the objective function will decrease because the factor $w_{jk}$ is higher than the other values used when updating the coordinates. 

		 This argument holds for any dimension $l$, any relation in $\mathcal{R}$, and for the categorical attributes represented as nodes in a bipartite graph.
        Hence, the value of the objective function will decrease at each iteration and when the decrement is smaller than $\epsilon$, the algorithm terminates.
\end{proof}

\textbf{Complexity analysis.} Lines $4-7$ in Algorithm~\ref{fig:algo} compute the contribution provided by relational information, its time complexity is $O(|R|\cdot m\cdot d)$, where $m$ is the number of edges. Lines $8-10$ compute the contribution given by the categorical data, whose time complexity is $O(C\cdot d)$. These operations are repeated for each node in the graph, thus, the time complexity of  lines $4-10$ for updating the object coordinates is $O(n(|R|\cdot m \cdot d+C\cdot d))$.
In line $11$ we apply the Gram-Schmidt orthonormalization algorithm with complexity $O(n\cdot d^2)$.
In lines $12-15$, the algorithm updates the coordinates of the categorical data objects, it requires $O(n\cdot C\cdot d)$. Finally, in lines $17-19$ we use K-Means algorithm with complexity $O(t\cdot d^2\cdot n)$ setting $K$ to $d-1$. We denote by $t$ the number of iterations of K-Means. The total time complexity of Algorithm~\ref{fig:algo} is $O(I(n\cdot|R|\cdot m \cdot d+n\cdot d^2+n\cdot C\cdot d)+t\cdot d^2\cdot n)$ where $I$ is the number of iterations the algorithm needs for converging.




\section{Experiments}\label{sec:exp}
In this section, we evaluate the performance of SpectralMix against a variety of state-of-the-art methods. Our focus is on node embedding and clustering tasks. Due to space limitations, we only can show a selection of the results. 

\textbf{Datasets.}
To compare SpectralMix with other baselines we use four datasets, which are multigraphs with node attributes, and two datasets, which are multigraphs without node attributes. To make fair comparisons with HAN \cite{han2019}, and DMGI~\cite{park2019unsupervised}, which are the most relevant baseline methods for attributed multigraphs, we evaluate our proposed algorithm on the same datasets used in~\cite{han2019} and in~\cite{park2019unsupervised}, i.e., ACM, and IMDB. Similarly, we use the datasets, DBLP and Flickr, that are used  for evaluating the performance of DMGC~\cite{luo2020deep}, which is the most relevant method for multi-relational graphs without node attributes.

\textbf{ACM}~\cite{park2019unsupervised}  is a multigraph with $2$ types of relations where nodes represent papers and edges are inferred via Authors collaborations (PAP) and Subjects of the paper (PSP). 
The task is to cluster the data into three clusters (Database, Wireless Communication, and Data Mining).

\textbf{IMDB}~\cite{park2019unsupervised}  is a multigraph with $2$ types of relations where nodes represent movies and edges are inferred via movie-actors (MAM) and movie-directors (MDM) relations. 
The task is to cluster the data into three clusters (Action, Comedy, Drama) according to movie genre.

\textbf{DBLP}~\cite{luo2020deep} is a multigraph of $4$ relations, that can be considered as $4$ graphs: 
a) collaboration graph – consists of edges between authors who have published a paper together, b) citation graph – consists of edges between authors, an edge represents a citation between authors, c) co-citation graph - consists of edges between papers, an edge represents that papers have cited the same source, and d) authorship graph – consists of edges of authors and papers, an edge represents an author who has published a paper. 
The task is to cluster the author and paper nodes into three cluster (AI, computer graphics, and computer networks) according to research areas.

\textbf{Flickr}~\cite{luo2020deep} is a dataset that does not have features and consists of two relations: a) friendship – represents an edge between users who are friends on social network, b) tag-similarity – represents an edge between users who have tag similarities among them. 
The task for this dataset is to cluster the user nodes into seven cluster, according to social groups.

\textbf{Brain Networks (BN)} can be defined as weighted graphs, where nodes are the brain regions,  edges are connections between regions, and weights denote distances between regions.
In our experiments, we use a publicly available dataset project~\cite{conf.fninf.2013} that contains data of $52$ Typically Developed (TD) children and $49$ children suffering from Autism Spectrum Disorder (ASD).
After preprocessing 
\cite{TZOURIOMAZOYER2002273}, we obtain, for each subject, a graph with $116$ nodes. We also use a second detaset~\cite{jakub_vohryzek_2020_3758534} that provides structural and functional brain networks from $27$ schizophrenic (SCZ) patients and $27$ matched healthy adults (HL).
The brain was divided into $83$ approximately equally sized regions.
For each node we are given the following attributes: average apparent diffusion coefficient, fiber density, generalized fractional anisotropy, length of fibers and number of fibers.
We perform two types of experiments. 
For the first dataset, in order to obtain a multi-relational graph, we aggregate the graphs of each subject in the groups ASD and TD. These multi-relational graphs are defined over the same set of nodes as the original graphs, while the weight and the type of relation of each edge are those of each subject in the original groups.
For the second dataset, we select at random a subject in each group, i.e., SCZ and HL, and we build a multi-relational graph where we have two edge types: the functional and structural connectivity measured with MRI and fMRI. We describe this results in details in Section~\ref{sec:brain}.

\textbf{Baseline methods.}
We compare SpectralMix with the state-of-the-art methods described in Section~\ref{sec:rel_works}, i.e., ANRL~\cite{ijcai2018-438},  DGI~\cite{velickovic2018deep}, DMGC~\cite{luo2020deep}, CrossMNA~\cite{10.1145/3308558.3313499},  HAN~\cite{han2019}, MARINE~\cite{10.1145/3308558.3313715}, and DMGI~\cite{park2019unsupervised}. We report their main characteristics in Table~\ref{table:comp}.

\begin{table}[t]
    \centering
    \caption{Properties of the compared methods}
    \label{table:comp}
    \begin{tabular}{ cc c c}
        \hline
            & Multigraphs  & Attribute & Unsupervised\\
        \hline
            ANRL & \color{red} \emph{X} & \color{blue} \checkmark & \color{blue} \checkmark\\
            DGI	& \color{red} \emph{X}  & \color{blue} \checkmark & \color{blue} \checkmark \\
        \hline
            DMGC & \color{blue} \checkmark & \color{red} \emph{X} & \color{blue} \checkmark \\
            CrossMNA	& \color{blue} \checkmark	& \color{red} \emph{X} & \color{blue} \checkmark \\
        \hline
            HAN & \color{blue} \checkmark	& \color{blue} \checkmark & \color{red} \emph{X} \\
            MARINE & \color{blue} \checkmark	& \color{blue} \checkmark & \color{blue} \checkmark \\
            DMGI	& \color{blue} \checkmark	& \color{blue} \checkmark & \color{blue} \checkmark \\
        \hline
            \textbf{SpectralMix}	& \color{blue} \checkmark	& \color{blue} \checkmark & \color{blue} \checkmark \\
        \hline
    \end{tabular}
\end{table}

 \begin{table*}[t]
\caption{Performance on node clustering on real-world datasets (\textit{NA}: Not Applicable).}
\label{table:res}
\begin{center}
\begin{tabular}{ c|c c|c c|c c|c c|c c|c c  } 
\hline
\multirow{2}{*}{ } & \multicolumn{2}{c|}{ACM}  & \multicolumn{2}{c|}{IMDB} & \multicolumn{2}{c|}{DBLP} & \multicolumn{2}{c|}{Flickr} & \multicolumn{2}{c|}{BrainNetwork-TD} & \multicolumn{2}{c}{BrainNetwork-ASD} \\  \cline{2-13} 
& NMI & ARI & NMI & ARI & NMI & ARI & NMI & ARI & NMI & ARI & NMI & ARI  \\
 \hline
 ANRL & 0.46 & 0.41  & 0.11  & 0.11 & 0.12 & 0.1 & 0.05 & 0.05 & NA & NA & NA & NA \\
 DGI & 0.63 & 0.61 & 0.15 & 0.14 & 0.08 & 0.05 & 0.09 & 0.09 & 0.13 & 0.04 & 0.13 & 0.04 \\
 \hline
 CrossMNA & 0.32 & 0.31 & 0.15 & 0.15 & 0.27 & 0.22 & 0.23 & 0.19 & NA & NA & NA & NA \\
 DMGC & 0.42 & 0.37 & 0.04 & 0.04 & 0.39 & 0.32 & 0.33 & 0.29 & NA & NA & NA & NA \\
 \hline
 HAN & 0.64 & 0.62 & 0.15 & 0.14 & 0.04 & 0.08 & 0.1 & 0.12 & 0.18 & 0.07 & 0.16 & 0.05 \\
 MARINE & 0.46 & 0.43 & 0.03 & 0.02 & 0.18 & 0.11 & 0.22 & 0.14 & NA & NA & NA & NA \\
 DMGI & 0.68 & 0.65 & 0.19 & 0.15 & 0.09 & 0.04 & 0.1 & 0.07 & 0.12 & 0.29 & 0.13 & 0.02\\
 \hline
 \hline
 SpectralMix & \textbf{0.71} & \textbf{0.77} & \textbf{0.21} & \textbf{0.17} & \textbf{0.47} & \textbf{0.34} & \textbf{0.64} & \textbf{0.51} & \textbf{0.7} & \textbf{0.62} & \textbf{0.6} & \textbf{0.51}  \\
 \hline
\end{tabular}
\end{center}
\end{table*}
 
For fair comparisons with methods \textit{not} designed for multigraphs, ANRL and DGI, we obtained the final node embedding matrix \textbf{Z} of a multi-relational graph by computing the average of the node embeddings obtained from each single graph, similarly as in ~\cite{park2019unsupervised}.
For the other two methods in the second category that ignore the node attributes, i.e., DMGC and CrossMNA, we concatenated the raw attribute matrix \textbf{X} to the learned node embeddings \textbf{Z} i.e., \textbf{Z} $\leftarrow$ [\textbf{Z};\textbf{X}], similarly as in ~\cite{park2019unsupervised}. 

We underline that SpectralMix can operate on multi-relational graphs with or without attributes and it obtains a joint embedding for all the relation types of the graphs. SpectralMix is applicable to multi-relational graphs but it is not limited to them because the learned node embeddings that we obtain for single graphs are the same as the final embeddings of Laplacian Eigenmaps. 

\textbf{Evaluation Metrics.} SpectralMix is an unsupervised method and does not need any labeled data for training. We focus on the task of node clustering, which is a classical task for unsupervised methods.  
We compute the most commonly used metrics to assess the quality of the clustering results~\cite{han2019}: Normalized Mutual Information (NMI)~\cite{nmi}, and Adjusted Rand Index (ARI) ~\cite{nmi}, where a value  equal to $1$ indicates a perfect clustering.

\textbf{Experimental setup.} 
The final node embeddings obtained from SpectralMix are suitable for spectral clustering, i.e. subsequent application of K-Means and for visualization. Therefore, we apply the K-Means on the final node embeddings to perform node clustering by setting K to the number of clusters of the dataset, and we use K-Means++ for initialization. Since the performance of K-Means is affected by initial centroids, we repeat the process for 100 times and report the average results. 
In the comparison with the alternative baselines, we keep the same setup and parameters that are made publicly available by the authors.

\textbf{Performance evaluation.} 
The experimental results of six public datasets are shown in Table~\ref{table:res}, the best results are marked in bold. It is easy to notice that SpectralMix outperforms all the alternative methods. Moreover, the performance of SpectralMix improves with the volume of data. As we can see from the results, we obtain the highest NMI and ARI values for the ACM dataset which is the graph with the largest number of edges and has $1,870$ attributes.  Thus, SpectralMix achieves state-of-the-art performance on ACM dataset, with $5\%$ performance lift in NMI, and $12\%$ in ARI, compared to the results from the competitors.  

SpectralMix outperforms other baselines on other real-world datasets too. For the IMDB dataset the  improvement of SpectralMix in terms of NMI is $11\%$ and ARI is $13\%$, and for multigraphs without attributes, DBLP and Flickr, the percentages of improvement in terms of NMI are $21\%$ and $94\%$, respectively, compared to the best performer from competitors for each dataset. 

Considering the results in Table~\ref{table:res}, we observe that the flaws of all baseline methods is that they perform well only for one category of graphs. The strength of SpectralMix is that it is applicable to different graph types: multi-relational graphs with and without attributes. Therefore, it outperforms the most recent and well known methods in both categories, respectively. Our advantage lies in the importance of considering separately the weights that each graph structure and each category of the node attributes have in our embedding. Also, SpectralMix performs well when applied to single graphs.

In particular, we observe that the DMGI~\cite{park2019unsupervised} performance drops when it is applied to multi-relational graphs without attributes, and this is due to the crucial role that node attributes play for DMGI.  
Similarly, the performance of all baselines which are designed for attributed graphs, such as ANRL~\cite{ijcai2018-438}, HAN ~\cite{han2019}, and MARINE~\cite{10.1145/3308558.3313715}, is poor on multi-relational graphs without attributes. Albeit, MARINE integrates the node attributes, we observe that very different node categories can impact its performance, which is the case for its performance drop on the IMDB dataset. Whereas, MARINE performance improves when applied to multi-relational graphs without attributes, and it occurs as MARINE is more focused on preserving link structures. As expected HAN~\cite{han2019} performs well on attributed multi-relational graphs and there is a drop in performance for the other two datasets, multi-relational graphs without attributes. The lack of node properties has affected the performance drop, since the node level mechanism of HAN uses information from both graph structure and node attributes.

Experimental results on brain networks, are presented for SpectralMix and only three competitors, DMGI, HAN, and DGI. For other methods, experiments could not be conducted as those methods are not able to process the data due to their limitations, such as not being able to process decimal or negative values for edge weights or node attributes. In case of brain networks, the gap in performance is even higher between SpectralMix and other methods, and one of the factors is that baseline methods have to train and validate their models in order to have a satisfiable model. Since for the brain networks the number of nodes and edges is very limited, it has an impact on the performance of these baselines methods.

\begin{figure}[b]
\small{
  \centering
  \begin{subfigure}[h!]{.49\linewidth}
     \includegraphics[scale=0.2]{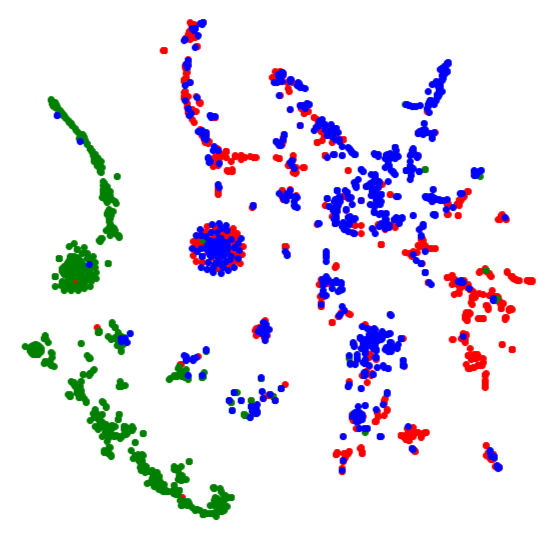}
    \caption{Without attributes.}\label{fig:withoutAtts}
  \end{subfigure}
  \begin{subfigure}[h!]{.49\linewidth}
     \centering\includegraphics[scale=0.2]{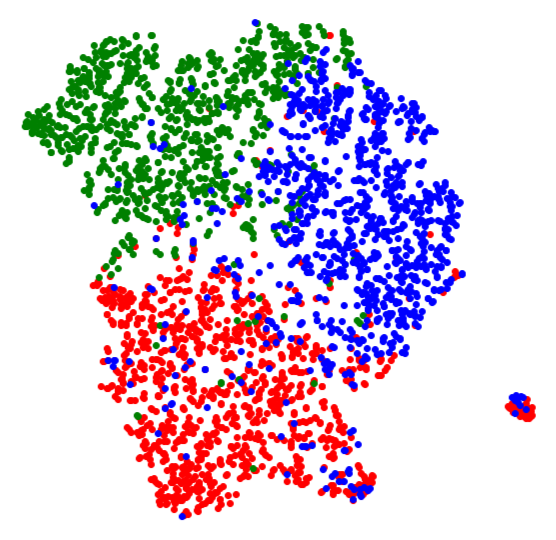}
    \caption{With attributes.}\label{fig:withAtts}
  \end{subfigure}
  
  \caption{Visualization of SpectralMix on ACM dataset.}
  \label{fig:visualizationSpectralMix}
  }
\end{figure}

\begin{figure*}[t]
\small{
  \centering

  \begin{subfigure}[t]{.245\linewidth}
    \centering\includegraphics[scale=0.2]{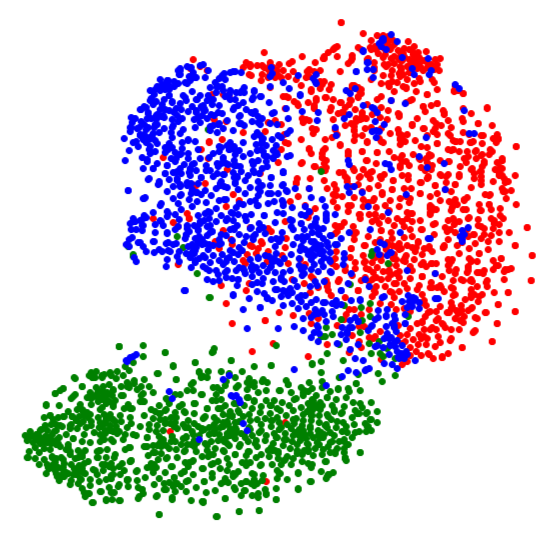}
    \caption{ANRL.}
  \end{subfigure}
  \begin{subfigure}[t]{.245\linewidth}
    \centering\includegraphics[scale=0.2]{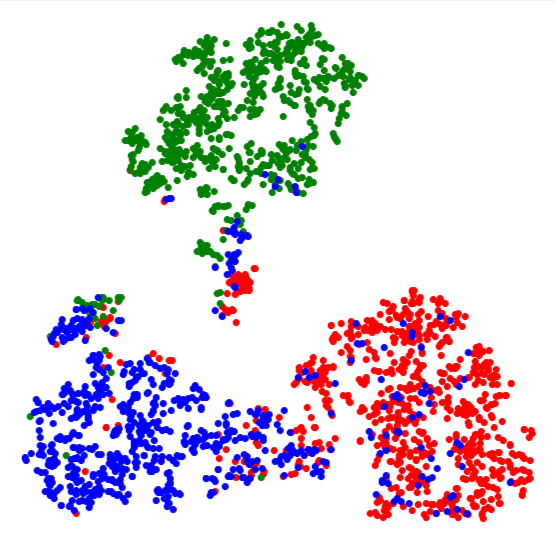}
    \caption{DGI.}
  \end{subfigure}
  \begin{subfigure}[t]{.245\linewidth}
    \centering\includegraphics[scale=0.2]{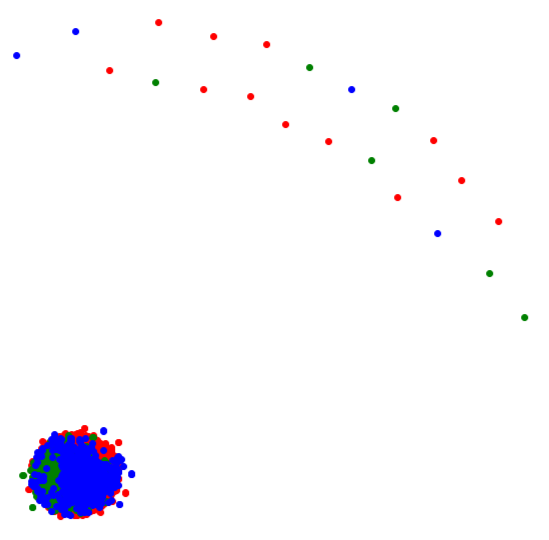}
    \caption{CrossMNA.}
  \end{subfigure}
  \begin{subfigure}[t]{.245\linewidth}
    \centering\includegraphics[scale=0.2]{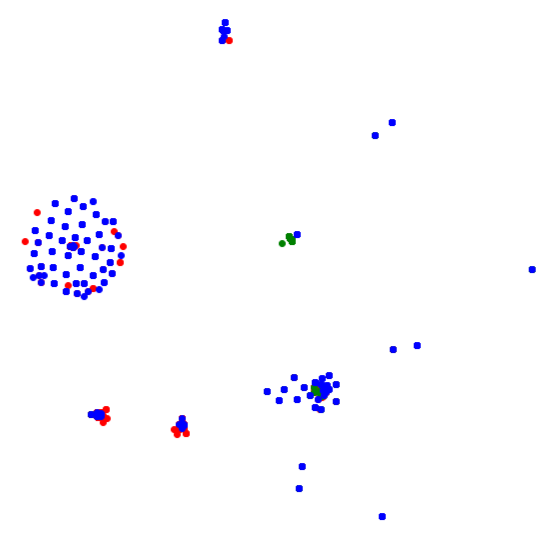}
    \caption{DMGC.}
  \end{subfigure}

  \medskip

  \begin{subfigure}[t]{.245\linewidth}
    \centering\includegraphics[scale=0.2]{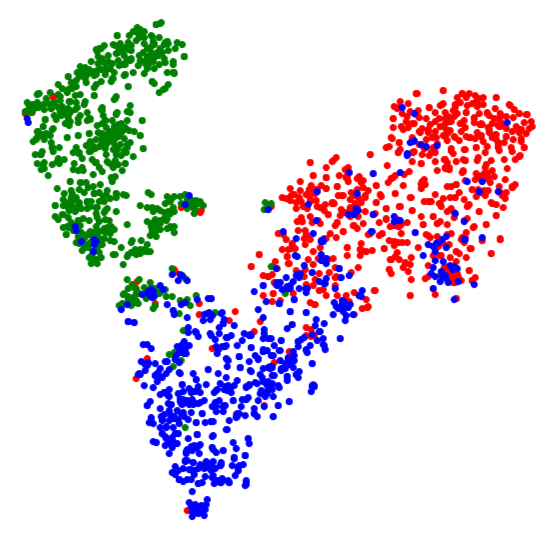}
    \caption{HAN.}
  \end{subfigure}
  \begin{subfigure}[t]{.245\linewidth}
    \centering\includegraphics[scale=0.2]{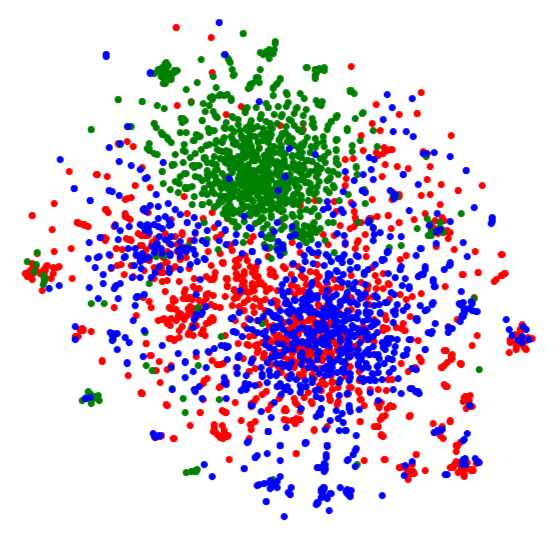}
    \caption{MARINE.}
  \end{subfigure}
  \begin{subfigure}[t]{.245\linewidth}
    \centering\includegraphics[scale=0.2]{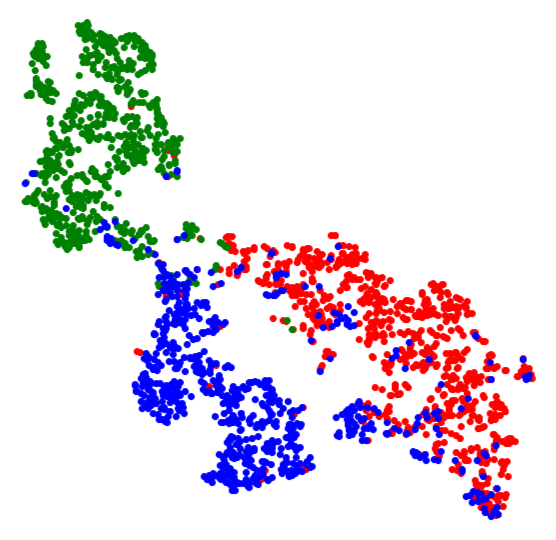}
    \caption{DMGI.}
  \end{subfigure}
  \begin{subfigure}[t]{.245\linewidth}
    \centering\includegraphics[scale=0.2]{img/SpectralMix.png}
    \caption{SpectralMix.}\label{fig:SpectralMixVis}
  \end{subfigure}
  
  \caption{Visualization on ACM dataset.}
  \label{fig:visualization}
  }
\end{figure*}

\textbf{Visualization.} To further demonstrate the effectiveness of SpectralMix over comparison methods, we show in Figure~\ref{fig:visualization} the visualization of all methods on ACM dataset.  
For the sake of fair comparison we apply visualization tool t-SNE ~\cite{tsne} on the node embeddings obtained from all methods. We set the same parameter values for visualization for all embeddings. The nodes are colored according to the research area: Database, Wireless Communication, and Data Mining.
From Figure~\ref{fig:visualization}, we can observe that ANRL, CrossMNA, DMGC, and MARINE do not perform well as the nodes belonging to different research areas are mixed with each other. Also, we can note that SpectralMix, DMGI, HAN, and DGI are able to better distinguish different types of nodes. It is clear that SpectralMix is able to separate the nodes in different research areas with more distinct boundaries and higher number of nodes clustered correctly. Additionally, it is noticeable on the SpectralMix node embedding in Figure~\ref{fig:SpectralMixVis} that there is a distinct group of nodes, on the right side, which represent the noise data or \textit{outliers}.
These nodes are only connected  with each other in PAP relation type of ACM dataset and do not share the same properties or attributes with other nodes.
Thus, we can see that SpectralMix is robust against outliers.
 
It is worth mentioning that the outlier nodes represent papers in both the types of relations in the graph of ACM, i.e., PAP and PSP. These papers have very few authors in common with other papers, i.e., they are connected be very few edges to other nodes in the PAP graph. Whereas, they are written by the same small group of authors, i.e., the edges between the outliers form a complete graph. 

We emphasize the importance of node attributes in the final node embeddings in Figure ~\ref{fig:visualizationSpectralMix}. We visualize the final node embeddings without node attributes in Figure~\ref{fig:withoutAtts} and with node attributes in Figure~\ref{fig:withAtts}. In Figure~\ref{fig:withoutAtts}  the cluster structure is not clear and nodes from two different clusters are closer in the lower dimensional space. Whereas, if we integrate the available information from node attributes we can improve our final node embeddings, Figure~\ref{fig:withAtts} yields the importance to integrate the node attributes in the final joint embedding, and this property of SpectralMix allows us to achieve a better performance on the clustering task.

\begin{figure}[b]
    \centering
    \includegraphics[scale=0.5]{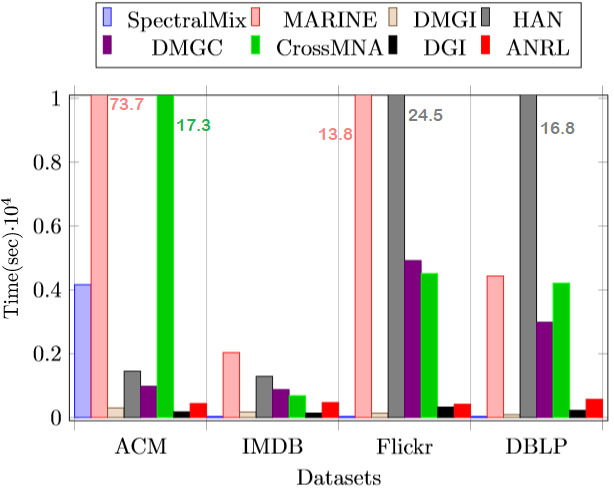}
    \vspace*{-2mm}
    \caption{Run time evaluation.}
    \label{fig:running_time}
\end{figure}

\textbf{Run time evaluation.} For 4 datasets analysed in Table~\ref{table:res}, we show the run time evaluation for all methods in Figure~\ref{fig:running_time}. We do not include the brain networks on the run time evaluation as we do not have results on these datasets for all the alternative baselines.
For the alternative baselines, we report the run time needed to converge.
It is noticeable that SpectralMix is faster than other methods on three datasets, i.e., IMDB, Flickr, and DBLP. In the ACM dataset, SpectralMix is slower than other methods and this occurs due to higher dimensionality of the output. Although, we recall that our algorithm is able to achieve better performance as showed in Table~\ref{table:res}.
Whereas, for other baseline methods, i.e. MARINE and CrossMNA, the main reason for their run time is the high number of edges in ACM dataset.
  
Our method, SpectralMix, outperforms also the alternative baselines on datasets without node attributes, Flickr, and DBLP.
The strength of SpectalMix is highlighted by these experiments. It outperforms baseline methods on different types of graphs: multi-relational graphs with node attributes and multi-relational graphs without node attributes.

\begin{figure}[h!]
\begin{center}
\begin{tikzpicture}[scale=0.44]
    \begin{axis}[
        xlabel={Iterations},
        ylabel={Objective Function ($\cdot 10^{4}$)},
        xmin=0, xmax=5000,
        ymin=3.715, ymax=3.736,
        xtick={1, 1000, 2000, 3000, 4000, 5000},
        ytick={ 3.715, 3.72,3.725,3.73,3.735},
        legend style ={ at={(1.03,1)}, 
            anchor=north west, draw=black, 
            fill=white,align=left},
        cycle list name=black white,
        ymajorgrids=false,
        ylabel near ticks,
        label style={font=\huge},
        tick label style={font=\huge},
    ]
    \addplot[label=l1,
        color=blue,
        mark=triangle*
        ]
        coordinates {
        (1,3.7323)(100,3.7192)(200,3.7192)(500,3.7192)(1000,3.7192)(2000,3.7192)(3000,3.7192)(4000,3.7192)(5000,3.7192)
        };
    \end{axis}
    \end{tikzpicture}
    \begin{tikzpicture}[scale=0.44]
    \begin{axis}[
        xlabel={Iterations},
        ylabel={Objective Function ($\cdot 10^{5}$)},
        xmin=0, xmax=5000,
        xtick={1, 1000, 2000, 3000, 4000, 5000},
        ymin=109.08500, ymax=109.20000,
        ytick={109.08500, 109.11500, 109.13500, 109.15500, 109.17500, 109.19500},
        ymajorgrids=false,legend style={at={(0.5,1.1)},anchor=north},
        ylabel near ticks,
        label style={font=\huge},
        tick label style={font=\huge}
    ]
    \addplot[label=l1,
        color=red,
        mark=oplus*,
        ]
        coordinates {
        (1,109.1816829)(100,109.0927661)(200,109.0927661)(500,109.0927661)(1000,109.0927661)(2000,109.0927661)(3000,109.0927661)(4000,109.0927661)(5000,109.0927661)
        };
    \end{axis}
    \end{tikzpicture}
  \end{center}
\caption{Objective function convergence on two real-world datasets: IMDB (left) and ACM (right).}
 \label{fig:ObjectiveFunction}
\end{figure}
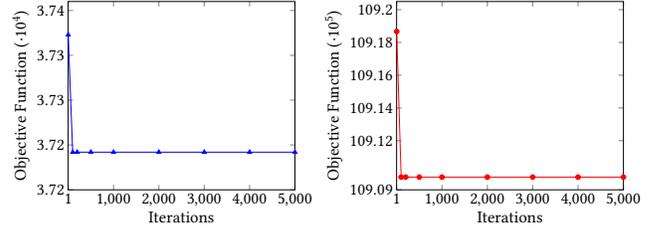

\textbf{Convergence.} In Section~\ref{sec:emb} we state that the value of our objective function decreases and when the difference in the value $\mathcal{F}$ between two consecutive iterations is smaller than $\epsilon$, our algorithm converges. We denote the number of these iterations by $I$ and we show experimentally that, in practice, SpectralMix converges with few iterations. For the two largest datasets, ACM and IMDB, $I$ is equal to $20$ and $60$ respectively, for $\epsilon=0.0001$. In Figure~\ref{fig:ObjectiveFunction} we can see that the value $\mathcal{F}$ of the objective function drops and remains almost constant even if we let the algorithm run for $5000$ iterations. 
Similarly, the performance of SpectralMix measured by the values of NMI when $I$ ranges in $[1, 5000]$ remain constant after reaching convergence.

\begin{figure}[b]
  \centering
   \includegraphics[scale=0.35]{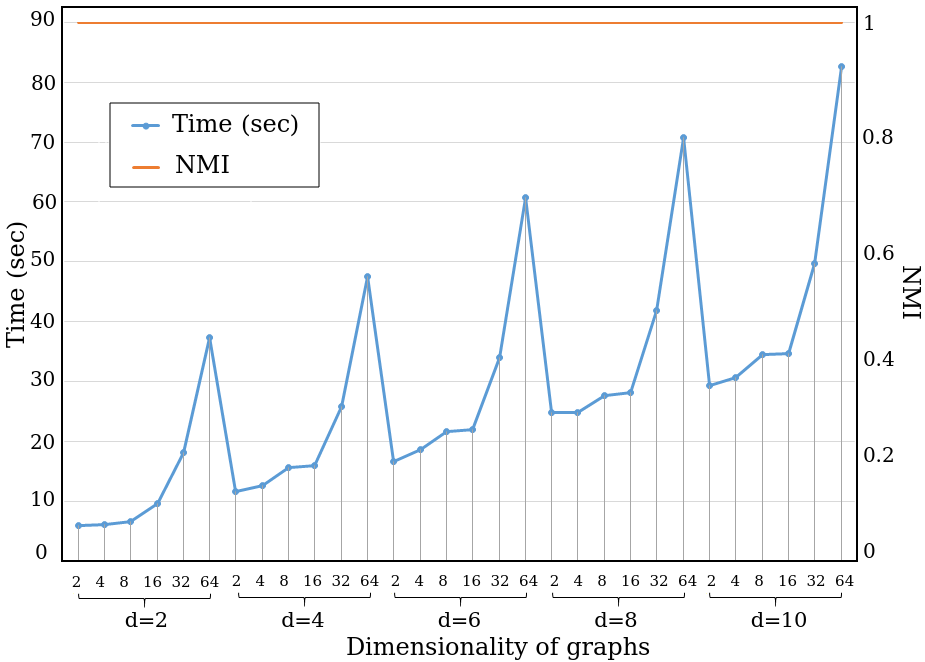}	
	\caption{Performance analysis on synthetic data.}\label{fig:dim_emb}
\end{figure}

\textbf{Parameter Analysis.} As we have mentioned in Section $3$, there is only one parameter $d$, desired dimensionality, that can be tuned for SpectralMix\footnote{The values selected for $d$ are reported in the Appendix.}. To show the impact of dimensionality on SpectralMix performance, we use variations of synthetic datasets, differing on number of relations between nodes. Figure~\ref{fig:dim_emb} shows the NMI values and run time of SpectralMix on synthetics dataset, with different dimensional embeddings as outputs. On these variations of synthetic datasets the NMI value is $1$ for all different dimensional embeddings with $d$ in the range $[2,64]$, and is not affected by different numbers of relations between nodes starting from $2$ up to $10$. We note that the running time increases slightly by having more relations between nodes and higher embeddings, but this is a common feature for methods that use all the available information for computations as the complexity increases too. 

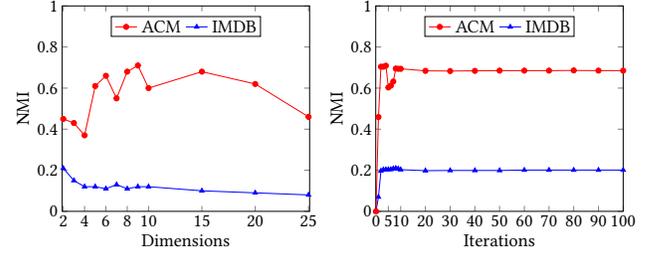
\begin{figure}[t]
\begin{center}
\begin{tikzpicture}[scale=0.48]
    \begin{axis}[
    xlabel={Dimensions},
    ylabel={NMI},
    xmin=1.9, xmax=25.1,
    ymin=0.0, ymax=1.0,
    ytick={0, 0.2 ,0.4,0.6,0.8,1},
    xtick={2,4,6,8,10,15,20,25},
    yticklabels={0, 0.2 ,0.4,0.6,0.8,1},
    ylabel near ticks,
    ymajorgrids=false,legend style={at={(0.5,1.1)},anchor=north},
    label style={font=\huge},
    tick label style={font=\huge}
]
\addplot[label=l1,
    color=red,
    mark=oplus*,
    ]
    coordinates {(2,0.45)(3,0.43)(4,0.37)(5,0.61)(6,0.66)(7,0.55)(8,0.68)(9,0.71)(10,0.60)(15,0.68)(20,0.62)(25,0.46)
    };
    \label{x21}
\addplot[
    color=blue,
    mark=triangle*,
    ]
     coordinates {(2,0.21)(3,0.15)(4,0.12)(5,0.12)(6,0.11)(7,0.13)(8,0.11)(9,0.12)(10,0.12)(15,0.10)(20,0.09)
(25,0.08)};\label{x31}
\node [draw,fill=white, style={font=\huge}] at (rel axis cs: 0.5,0.9) {\shortstack[l]{
\ref{x21} ACM
\ref{x31} IMDB}}; 
\end{axis}
\end{tikzpicture}
\begin{tikzpicture}[scale=0.48]
    \begin{axis}[
    xlabel={Iterations},
    ylabel={NMI},
    xmin=-0.1, xmax=100,
    ymin=0.0, ymax=1,
    ytick={0, 0.2,0.4,0.6,0.8,1},
    xtick={0,5,10,20,30,40,50,60,70,80,90,100},
    yticklabels={0, 0.2,0.4,0.6,0.8,1},
    ylabel near ticks,
    ymajorgrids=false,legend style={at={(0.5,1.1)},anchor=north},
    label style={font=\huge},
    tick label style={font=\huge}
    ]
    \addplot[label=l1,color=blue,mark=triangle*]
        coordinates {(0,0)(1,0.07)(2,0.197)(3,0.203)(4,0.203)(5,0.204)(6,0.204)(7,0.208)(8,0.21)(9,0.207)(10,0.203)(20,0.198)(30,0.199)(40,0.199)(50,0.199)(60,0.201)(70,0.201)(80,0.201)(90,0.201)(100,0.201)
        };\label{x3}
    \addplot[label=l1,color=red,mark=oplus*,]
    coordinates {(0,0)(1,	 0.459)(2,	 0.704)(3,	 0.704)(4,	 0.709)(5,	 0.603)(6,	 0.612)(7,	 0.632)(8,	 0.695)(9,	 0.694)(10, 0.694)(20, 0.684)(30, 0.683)(40, 0.684)(50, 0.685)(60, 0.685)(70, 0.685)(80, 0.686)(90, 0.685)(100,0.685)
    };\label{x2}
    \node [draw,fill=white, style={font=\huge}] at (rel axis cs: 0.5,0.9) {\shortstack[l]{
        \ref{x2} ACM
        \ref{x3} IMDB
    }};
    \end{axis}
\end{tikzpicture}
\end{center}
\caption{Parameter analysis on real-world datasets.}
\label{fig:parameterAnalysis}
\end{figure}

Figure~\ref{fig:parameterAnalysis} shows the impact of dimensionality on clustering performance and number of iterations SpectralMix needs to converge for two real-world datasets. As the value of $d$ increases the drop on performance is noticeable on IMDB dataset. For the ACM dataset we see that the performance changes over different dimensions and the best performance is achieved when we set dimensionality $d=9$. The main reasons are that: ACM is more dense and, PAP and PSP, have different number of edges ($29,281$ and $2,210,761$ respectively). Thus, this causes the runtime for ACM dataset to be higher than for other datasets, as shown in Figure~\ref{fig:running_time}.

\textbf{Ablation study}. To measure the impact of each component of SpectralMix, we conduct ablation studies on the largest dataset, ACM, and we report the results in Table ~\ref{table:compACM-SpectralMix}. We observe the following facts: 1) As expected, the node attributes help to improve the node clustering performance, therefore, node attributes have significant role for representation learning of nodes. 2) If we set the same weight factors  for all attribute categories then the impact is very high, as we see form the drop on the performance on clustering. 3) If we set the same weight factor for all different relation types, then the performance is affected but not in the same scale as it is for attribute categories.  4) The second summand, attribute contribution, in Eq.~\ref{eq:obj} indeed plays a significant role in optimization. Thus, the ablation study demonstrates that by assigning the different weights to relation types and categories of node attributes, SpectralMix can learn a meaningful node embedding. 

\begin{table}[b]
    \centering
    \caption{Result for ablation studies of SpectralMix.}
\label{table:compACM-SpectralMix}
\begin{tabular}{c c| c | c } 
\hline
\multicolumn{2}{c|}{ACM Dataset} & NMI  & ARI  \\ 
 \hline
 \multicolumn{2}{c|}{SpectralMix} & 0.71 & 0.77 \\
 \hline

\multicolumn{2}{l|}{1) No attributes} & 0.51 & 0.42 \\
\multicolumn{2}{l|}{2) Attributes Contribution: $ \alpha_1 = \dots = \alpha_\mathcal{|A|} $} & 0.01 & 0 \\
\multicolumn{2}{l|}{3) Relations Contribution: $ \alpha_1 = \dots = \alpha_{\mathcal{|R|}}$} & 0.55 & 0.61  \\
\hline
\multicolumn{1}{l|}{\multirow{2}{*}{4) Objective Function:}} & $\alpha_{\mathcal{R}} > \alpha_\mathcal{A} $ & 0.65 & 0.71 \\
\multicolumn{1}{l|}{}&  $\alpha_{\mathcal{R}} < \alpha_\mathcal{A} $ & 0.69 & 0.75 \\
\hline
\end{tabular}
\end{table}

\begin{figure}[t]
		\includegraphics[scale=0.095]{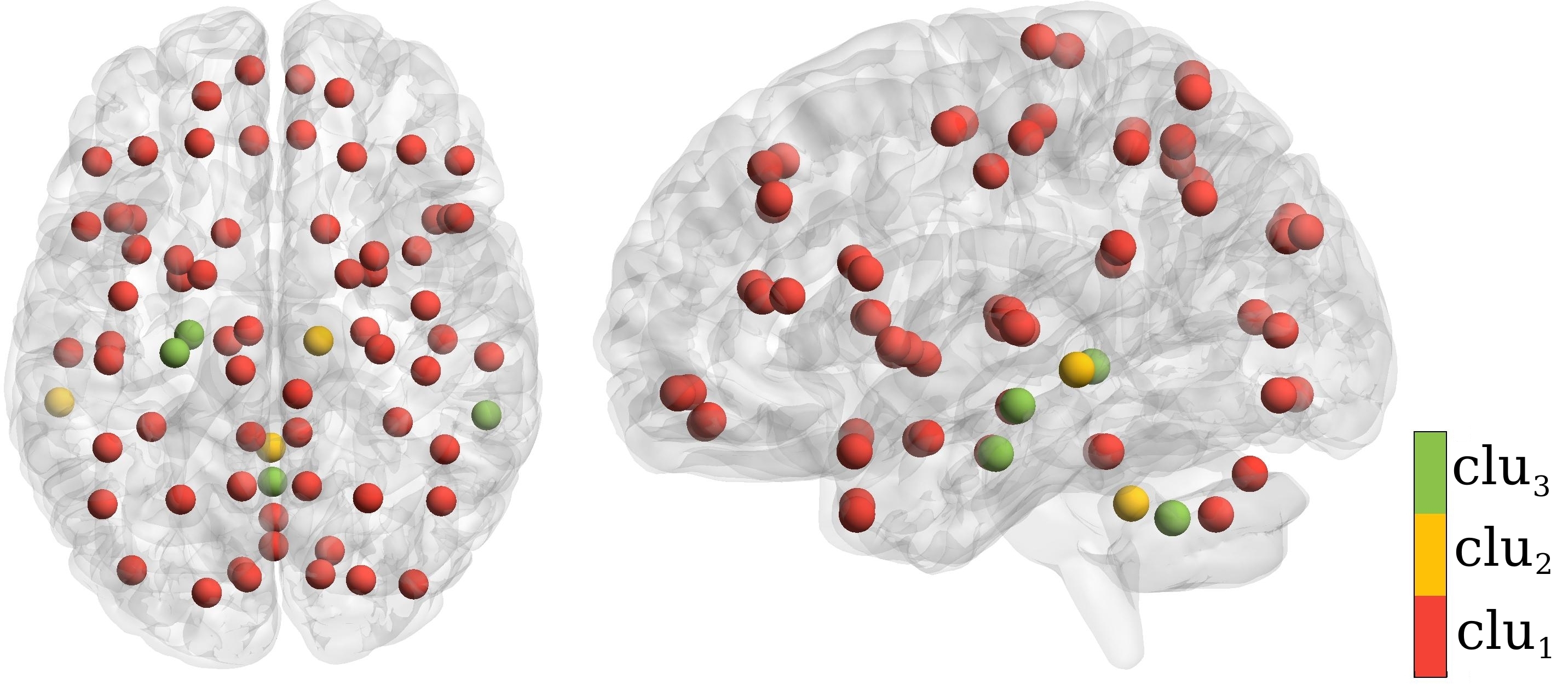}\\
		\includegraphics[scale=0.095]{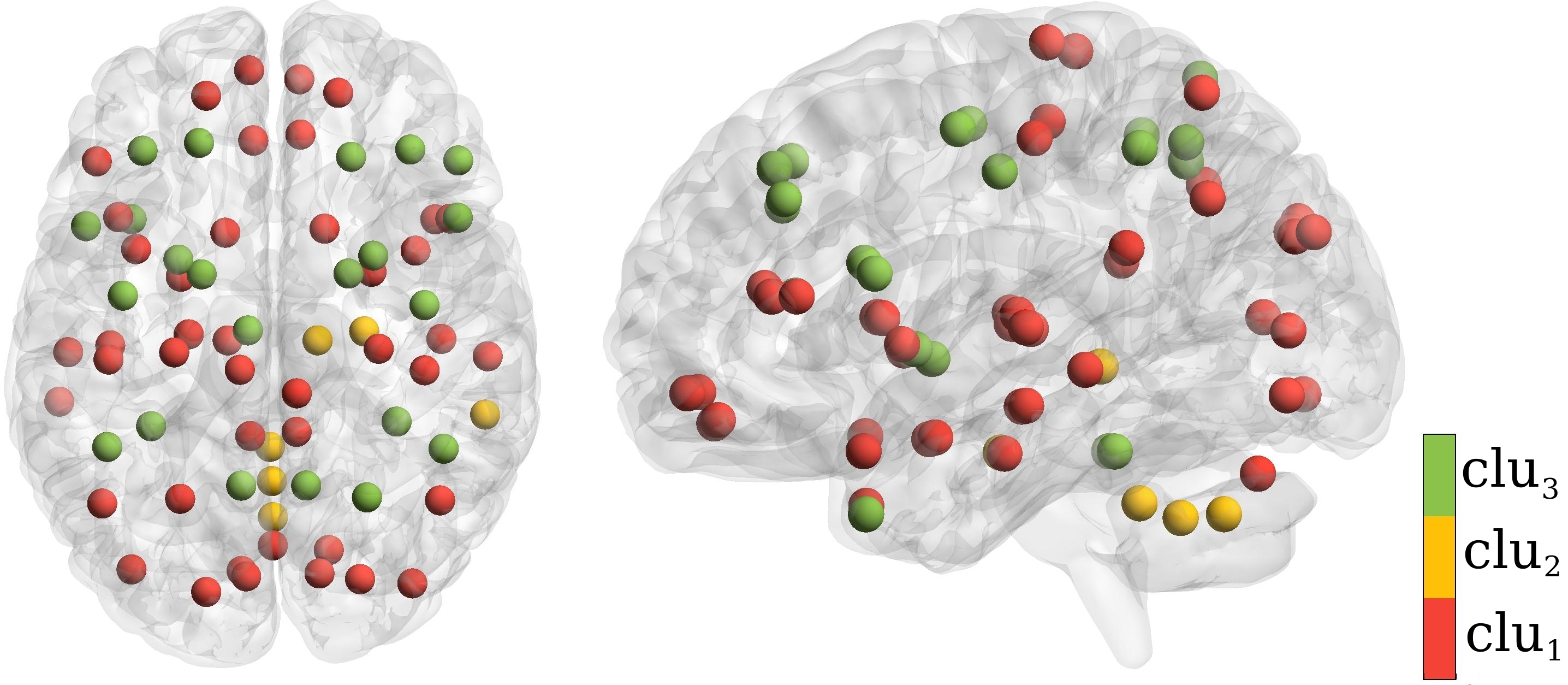}
	\caption{\small{Clustering for the graph of a subject in the HL group (top) and for the graph of a subject in the SCZ group (bottom).}}\label{fig:case_brain}
\end{figure} 

\textbf{Case Study: Brain Network}.\label{sec:brain}
In Figure~\ref{fig:case_brain} we plot the results obtained with SpectralMix. We plot the clusters obtained for an healthy subject $\mathcal{G}_{HL}$ (top) and those obtained for a subject $\mathcal{G}_{SCZ}$ affected by schizophrenia (bottom) in an axial and sagittal view of the brain.
We run experiments considering different number of clusters in the range $[2, 6]$, and we present the result with the highest NCut~\cite{NCut} value, therefore we plot the results for three clusters.
We first notice that in $\mathcal{G}_{HL}$ there is a large cluster, i.e., $clu_1$, while in $\mathcal{G}_{SCZ}$ many regions in the temporal and frontal lobe belong to a different cluster, i.e., $clu_3$. These findings are consistent with works in neuroscience~\cite{neuro1, neuro2} affirming that some specific cognitive deficits present in schizophrenia can be attributed to  parietal and frontal lobe dysfunction. Moreover other neuroimaging studies~\cite{neuro3} have provided evidence that schizophrenia is associated with reductions in the volume of tissue in the cerebrum and cerebellum, and a disruption of the neural connectivity between them. These results, are coherent with the clustering found by SpectarlMix where nodes in the cerebellar area belong to the same cluster in $\mathcal{G}_{SCZ}$ as they are characterized by reduced cerebral volume while in the healthy subject, the same nodes are assigned to $3$ different clusters.

\section{Future Work}

There are different open problems worth investigating. 
For the interpretation of the algorithm, it is interesting to quantify how much of the link information of each modality is preserved in each Eigenvector. Furthermore, we are interested in answering the question: Are there any dense sub-graphs of a certain link type which are associated with a certain subspace of the attributes?
Moreover, we can extend SpectralMix to perform other data mining tasks such as link prediction, node classification, and to infer labels of unlabeled nodes in sparsely labeled multi-graphs, extending the semi-supervised learning method in~\cite{DBLP:conf/kdd/YeZMPB17}.

\bibliographystyle{ACM-Reference-Format}
\bibliography{sample-base}
\newpage
\section*{APPENDIX: IMPLEMENTATION AND REPRODUCIBILITY}

The source code of SpectralMix is publicly available \footnote{\hyperlink{}{https://gitlab.cs.univie.ac.at/yllis19cs/spectralmixpublic}}. 

Our source code is implemented in Java and Python. We use Java to implement the main part of the SpectralMix algorithm and we have  used Python to run K-Means on node embeddings obtained from SpectralMix and other methods too. 
To run the code call: \emph{java jar -SpectralMix.jar} with the following argument \emph{source file name}, which is the name of the Matlab file where all the information about adjacency matrix (matrices) and node attributes are stored. Example: \emph{java -jar SpectralMix.jar flickr}.

To change parameters about desired dimensionality or number of iterations for each dataset, please use the class \emph{Sociopatterns.java} (package \emph{datautil}). 

Also, to conduct experiments on synthetic datasets and to generate synthetic datasets please use the class \emph{DataGeneratorMain.java}.

The datasets are stored in the folder named \emph{data}. Embeddings for each dataset are stored in the folder named \emph{embeddings}.

First we run the Java source code implementation, then we get the final embeddings, and we need to run Python script for K-Means on the final embeddings to get the cluster labels for each node, for this purpose we run the Python script stored on K-Means python notebook.

All experimental data are open source. Table ~\ref{table:stats}  summarizes their characteristics. 
A brief description for the construction of the first four datasets (\emph{ACM}, \emph{IMDB}, \emph{DBLP}, \emph{Flickr}) is provided below.

\begin{itemize}
    
    \item \emph{ACM}~\cite{park2019unsupervised}: The heterogeneous graph is constructed by using information from $3,025$ papers (P), $5,835$ authors (A) and $56$ subjects (S). There are $1870$ attributes for each paper.

    \item \emph{IMDB}~\cite{park2019unsupervised}: The constructed heterogeneous graph includes information from $3,550$ movies (M), $4,441$ actors (A), and $1,726$ directors (D). There are $2000$ attributes for each movie.

    \item \emph{DBLP}\cite{luo2020deep}: The collaboration graph has $2,401$ author nodes and $8,703$ edges. The citation graph has $6,000$ paper nodes and $10,003$ edges. The co-citation graph has $6,000$ paper nodes and $141,996$ edges. The authorship graph has $64,096$ edges. 

    \item \emph{Flickr}~\cite{luo2020deep}: Each relation type involves $10,364$ users as nodes. The friendship relation type has 401,302 edges. The tag-similarity relation type has 125,547 edges, where each edge represents the tag similarities between two users. Two nodes in these two relations(graphs) are linked if they refer to the same user. 
    
\end{itemize}

\begin{table}[h]
    \caption{Statistics of datasets.}
    \label{table:stats}
    \begin{center}
        \begin{tabular}{ lccccc}
        \hline
            datasets & \#relations & \#nodes & \#edges & \#atts & \#clusters\\
        \hline
            ACM	& 2	& 8,916 & 2,240,042 &	1,870 &  3\\
            IMDB & 2	& 9,717 & 80,216 &	2,000 & 3\\
            DBLP & 4	& 14,401 & 224,798 & 0 & 3\\
            Flickr	& 2	& 20,728 & 537,213 & 0 & 7\\
            BN-TD	& 51	& 116 & 66,744 & 6 & 3\\
            BN-ASD	& 49	& 116 & 62,907 & 6 & 3\\
            \hline
        \end{tabular}
    \end{center}
\end{table}

\subsection*{Experimental Setup: Parameter Settings}
We summarize details about the  parameter settings of SpectralMix. The only tuning parameter is $d$ and here we provide the desired dimensionality for each dataset:
\begin{itemize}
    
    \item \emph{ACM}: Dimensionality = 9. 
    
    \item \emph{IMDB}: Dimensionality = 2. 
    
    \item \emph{Flickr}: Dimensionality = 11. 
    
    \item \emph{DBLP}: Dimensionality = 2. 
    
    \item \emph{BrainNet-Healthy}: Dimensionality = 2. 
    
    \item \emph{BrainNet-Autism}: Dimensionality = 2.

\end{itemize}

Moreover, we set the number of iterations for convergence to $100$ for all datasets. 
In the experiments, we notice that the algorithm converges with less than $100$ iterations.

More details on how to run experiments are available at the provided link on the paper. 

For the comparison methods, we use the source code and the parameter settings provided by authors of each method and can be found in the following links:

\begin{itemize}
  
  \item \emph{ANRL~\cite{ijcai2018-438}}: python implemenation is publicly available at \url{https://github.com/cszhangzhen/ANRL}

  \item \emph{DGI~\cite{velickovic2018deep}}: python implementation is publicly available at \url{https://github.com/PetarV-/DGI}

  \item \emph{CrossMNA~\cite{10.1145/3308558.3313499}}: python implementation is publicly available at \url{https://github.com/ChuXiaokai/CrossMNA}

  \item \emph{DMGC~\cite{luo2020deep}}: python implementation is publicly available at \url{https://github.com/flyingdoog/DMGC}

  \item \emph{HAN~\cite{han2019}}: python implementation is publicly available at \url{https://github.com/Jhy1993/HAN}

  \item \emph{MARINE~\cite{10.1145/3308558.3313715}}: python implementation is publicly available at \url{https://github.com/ntumslab/MARINE}

  \item \emph{DMGI~\cite{park2019unsupervised}}: python implementation is publicly available at \url{https://github.com/pcy1302/DMGI}

\end{itemize}

\end{document}

%% file: img/fig1.tex
\begin{tikzpicture}[font=\huge,-,>=stealth',shorten >=1pt,auto,node distance=3cm,
  thick,main node/.style={circle,draw}, node/.style={sloped,anchor=south,auto=false}]
  \tikzset{edge/.style = {- = latex'}}

\node (1) at (-3.55, 2) {};
  \node[main node, minimum size=0.6cm] (1) at (-3.55,2) {};

\node (2) at (-5, 3) {};
  \node[main node, minimum size=0.6cm] (2) at (-5,3) {};
\node (3) at (-1, 3.5) {};
  \node[main node, minimum size=0.6cm] (3) at (-1,3.5) {};
\node (4) at (-3.25, 4) {};
 \node[main node, minimum size=0.6cm] (4) at (-3.25,4) {};

\node (7) at (-1.75, 2) {};
  \node[main node, minimum size=0.6cm] (7) at (-1.75,2) {};


\draw[edge] (1) to node[anchor=south ] { } (7);
\draw[edge] (4) to node[anchor=north] {} (7);
\draw[edge] (7) to node[anchor= east] { } (3);
\draw[edge] (2) to node[anchor=east] {} (4);


\node (1a) at (4, 2) {};
 \node[main node, minimum size=0.6cm] (1a) at (4,2) {};

\node (2a) at (2.5, 3) {};
  \node[main node, minimum size=0.6cm] (2a) at (2.5,3) {};
\node (3a) at (6.5, 3.5) {};
  \node[main node, minimum size=0.6cm] (3a) at (6.5,3.5) {};
\node (4a) at (4.25, 4) {};
 \node[main node, minimum size=0.6cm] (4a) at (4.25,4) {};

\node (7a) at (6, 2) {};
  \node[main node, minimum size=0.6cm] (7a) at (6,2) {};

\tikzset{box1/.style={draw=black, thick, rectangle, minimum height=0.3cm, minimum width=0.3cm}}

\node[box1, fill=pink] (c3)at (3.4, 1.2)  {};
\node[box1, fill=white] (c0)at (3.7, 1.2)  {};
\node[box1, fill=red!60] (c1)at (4, 1.2) {};
\node[box1, fill=green] (c2)at (4.3, 1.2)  {};
\node[box1, fill=cyan] (c3)at (2.8, 1.2)  {};
\node[box1, fill=yellow] (c3)at (3.1, 1.2)  {};

\node[box1, fill=pink] (c3)at (5.6, 1.3)  {};
\node[box1, fill=cyan] (c0)at (5.9, 1.3)  {};
\node[box1, fill=red!60] (c1)at (6.2, 1.3) {};
\node[box1, fill=green] (c2)at (6.5, 1.3)  {};
\node[box1, fill=white] (c3)at (6.8, 1.3)  {};
\node[box1, fill=yellow] (c3)at (5.3, 1.3)  {};

\node[box1, fill=pink] (c3)at (0.4, 2.3)  {};
\node[box1, fill=white] (c0)at (0.7, 2.3)  {};
\node[box1, fill=red!60] (c1)at (1, 2.3) {};
\node[box1, fill=green] (c2)at (1.3, 2.3)  {};
\node[box1, fill=cyan] (c3)at (1.6, 2.3)  {};
\node[box1, fill=yellow] (c3)at (1.9, 2.3)  {};

\node[box1, fill=pink] (c3)at (3.4, 4.6)  {};
\node[box1, fill=white] (c0)at (3.7, 4.6)  {};
\node[box1, fill=red!60] (c1)at (4, 4.6) {};
\node[box1, fill=cyan] (c2)at (4.3, 4.6)  {};
\node[box1, fill=green] (c3)at (2.8, 4.6)  {};
\node[box1, fill=yellow] (c3)at (3.1, 4.6)  {};

\node[box1, fill=pink] (c3)at (5.8, 4.3)  {};
\node[box1, fill=white] (c0)at (6.1, 4.3)  {};
\node[box1, fill=red!60] (c1)at (6.4, 4.3) {};
\node[box1, fill=green] (c2)at (6.7, 4.3)  {};
\node[box1, fill=cyan] (c3)at (7, 4.3)  {};
\node[box1, fill=yellow] (c3)at (5.5, 4.3)  {};

\draw[edge] (1a) to node[anchor=south ] { } (7a);
\draw[edge] (4a) to node[anchor=north] {} (7a);
\draw[edge] (7a) to node[anchor= east] { } (3a);
\draw[edge] (2a) to node[anchor=east] {} (4a);

\draw[bend left,dashed] (2a) to node[] {} (4a);
\draw[dashed] (1a) to node[] {} (2a);
\draw[bend left,dashed] (7a) to node[] {} (3a);

\draw[dotted] (1a) to node[] {} (4a);
\draw[dotted, bend left] (1a) to node[] {} (2a);
\draw[bend left,dotted] (4a) to node[] {} (7a);
\end{tikzpicture}

%% file: img/fig2.tex
\begin{tikzpicture}[font=\huge,-,>=stealth',shorten >=1pt,auto,node distance=3cm,
  thick,main node/.style={circle,draw}, node/.style={sloped,anchor=south,auto=false}]
  \tikzset{edge/.style = {- = latex'}}

\node (1) at (10, 1) {};
  \node[main node, minimum size=0.6cm] (1) at (10,1) {};
\node (2) at (10, 3) {};
  \node[main node, minimum size=0.6cm] (2) at (10,3) {};
\node (3) at (10, 4) {$v_j$};
  \node[main node, minimum size=0.6cm] (3) at (10,4) {};
\node (4) at (10, 2) {$v_i$};
 \node[main node, minimum size=0.6cm] (4) at (10, 2) {};
\node (7) at (10, 5) {};
  \node[main node, minimum size=0.6cm] (7) at (10 ,5) {};

\node (2a) at (12, 3) {};
  \node[main node, minimum size=0.6cm] (2a) at (12,3) {};
\node (3a) at (12, 4) {};
  \node[main node, minimum size=0.6cm] (3a) at (12,4) {};
\node (4a) at (12, 2) {};
 \node[main node, minimum size=0.6cm] (4a) at (12, 2) {};

\node(5) at (9.6, 6) { Nodes};
  \node(5) at (12.5,6) {Categories of};
\node(5) at (12.5, 5.6) { attribute };
\node(5) at (12.5, 5.1) {  "Hometown"};

  \node(5) at (13.1,4) {Vienna};
\node(5) at (13, 3) { Rome };
\node(5) at (13, 2) { Paris};

\draw[edge] (1) to node[anchor=south ] { } (4a);
\draw[edge] (4) to node[anchor=north] {} (2a);
\draw[edge] (2) to node[anchor= east] { } (4a);
\draw[edge] (7) to node[anchor=east] {} (3a);
\draw[edge] (3) to node[anchor=east] {} (3a);


\node (1a) at (4, 2) {};
 \node[main node, minimum size=0.6cm] (1a) at (4,2) {};

\node (2a) at (2.5, 3) {};
  \node[main node, minimum size=0.6cm] (2a) at (2.5,3) {};
\node (3a) at (6.5, 3.5) {$v_i$};
  \node[main node, minimum size=0.6cm] (3a) at (6.5,3.5) {};
\node (4a) at (4.25, 4) {$v_j$};
 \node[main node, minimum size=0.6cm] (4a) at (4.25,4) {};

\node (7a) at (6, 2) {};
  \node[main node, minimum size=0.6cm] (7a) at (6,2) {};

\node(5) at (5, 5.5) { $\{...,$ Hometown$,...\}$};

\tikzset{box1/.style={draw=black, thick, rectangle, minimum height=0.3cm, minimum width=0.3cm}}

\node[box1, fill=pink] (c3)at (3.4, 1.2)  {};
\node[box1, fill=white] (c0)at (3.7, 1.2)  {};
\node[box1, fill=red!60] (c1)at (4, 1.2) {};
\node[box1, fill=green] (c2)at (4.3, 1.2)  {};
\node[box1, fill=cyan] (c3)at (2.8, 1.2)  {};
\node[box1, fill=yellow] (c3)at (3.1, 1.2)  {};

\node[box1, fill=pink] (c3)at (5.6, 1.3)  {};
\node[box1, fill=cyan] (c0)at (5.9, 1.3)  {};
\node[box1, fill=red!60] (c1)at (6.2, 1.3) {};
\node[box1, fill=green] (c2)at (6.5, 1.3)  {};
\node[box1, fill=white] (c3)at (6.8, 1.3)  {};
\node[box1, fill=yellow] (c3)at (5.3, 1.3)  {};

\node[box1, fill=pink] (c3)at (0.7, 2.3)  {};
\node[box1, fill=white] (c0)at (1, 2.3)  {};
\node[box1, fill=red!60] (c1)at (1.3, 2.3) {};
\node[box1, fill=green] (c2)at (1.6, 2.3)  {};
\node[box1, fill=cyan] (c3)at (1.9, 2.3)  {};
\node[box1, fill=yellow] (c3)at (2.2, 2.3)  {};

\node[box1, fill=pink] (c3)at (2.3, 4.6)  {Vienna};
\node[box1, fill=white] (c0)at (3.3, 4.6)  {};
\node[box1, fill=red!60] (c1)at (3.6, 4.6) {};
\node[box1, fill=cyan] (c2)at (3.9, 4.6)  {};
\node[box1, fill=green] (c3)at (1, 4.6)  {};
\node[box1, fill=yellow] (c3)at (1.3, 4.6)  {};

\node[box1, fill=pink] (c3)at (6.3, 4.3)  {};
\node[box1, fill=white] (c0)at (7.15, 4.3)  {Rome};
\node[box1, fill=red!60] (c1)at (8.0, 4.3) {};
\node[box1, fill=green] (c2)at (8.3, 4.3)  {};
\node[box1, fill=cyan] (c3)at (8.6, 4.3)  {};
\node[box1, fill=yellow] (c3)at (6, 4.3)  {};

\draw[edge] (1a) to node[anchor=south ] { } (7a);
\draw[edge] (4a) to node[anchor=north] {} (7a);
\draw[edge] (7a) to node[anchor= east] { } (3a);
\draw[edge] (2a) to node[anchor=east] {} (4a);

\draw[bend left,dashed] (2a) to node[] {} (4a);
\draw[dashed] (1a) to node[] {} (2a);
\draw[bend left,dashed] (7a) to node[] {} (3a);

\draw[dotted] (1a) to node[] {} (4a);
\draw[dotted, bend left] (1a) to node[] {} (2a);
\draw[bend left,dotted] (4a) to node[] {} (7a);
\end{tikzpicture}